%% file: main.tex
\documentclass[10pt,twocolumn,letterpaper]{article}

\usepackage{cvpr}              
\usepackage{booktabs}   
\usepackage{graphicx}   
\usepackage{siunitx}    

\input{preamble}

\definecolor{cvprblue}{rgb}{0.21,0.49,0.74}
\usepackage[pagebackref,breaklinks,colorlinks,citecolor=cvprblue]{hyperref}

\def\paperID{4592} 
\def\confName{CVPR}
\def\confYear{2026}

\title{\shortname: Autoregressive Video Generation via Next-Frame \& Scale Prediction}

\author{
Longbin Ji$^{*}$ \quad Xiaoxiong Liu$^{*}$ \quad Junyuan Shang$^{\dagger}$ \\
Shuohuan Wang \quad Yu Sun \quad Hua Wu \quad Haifeng Wang \vspace{1mm}\\
ERNIE Team, Baidu\\
{\tt\small \{jilongbin, liuxiaoxiong, shangjunyuan\}@baidu.com}\\
{\tt\small \{wangshuohuan, sunyu02, wu\_hua, wanghaifeng\}@baidu.com}\\
\small $^{*}$Equal contribution \quad $^{\dagger}$Project lead \\
\small \url{https://ernie-research.github.io/VideoAR/}
}

\begin{document}

\twocolumn[{
\renewcommand\twocolumn[1][]{#1}
\maketitle 
\begin{center}
\vspace{-6mm}
    \centering
    \includegraphics[width=1\textwidth]{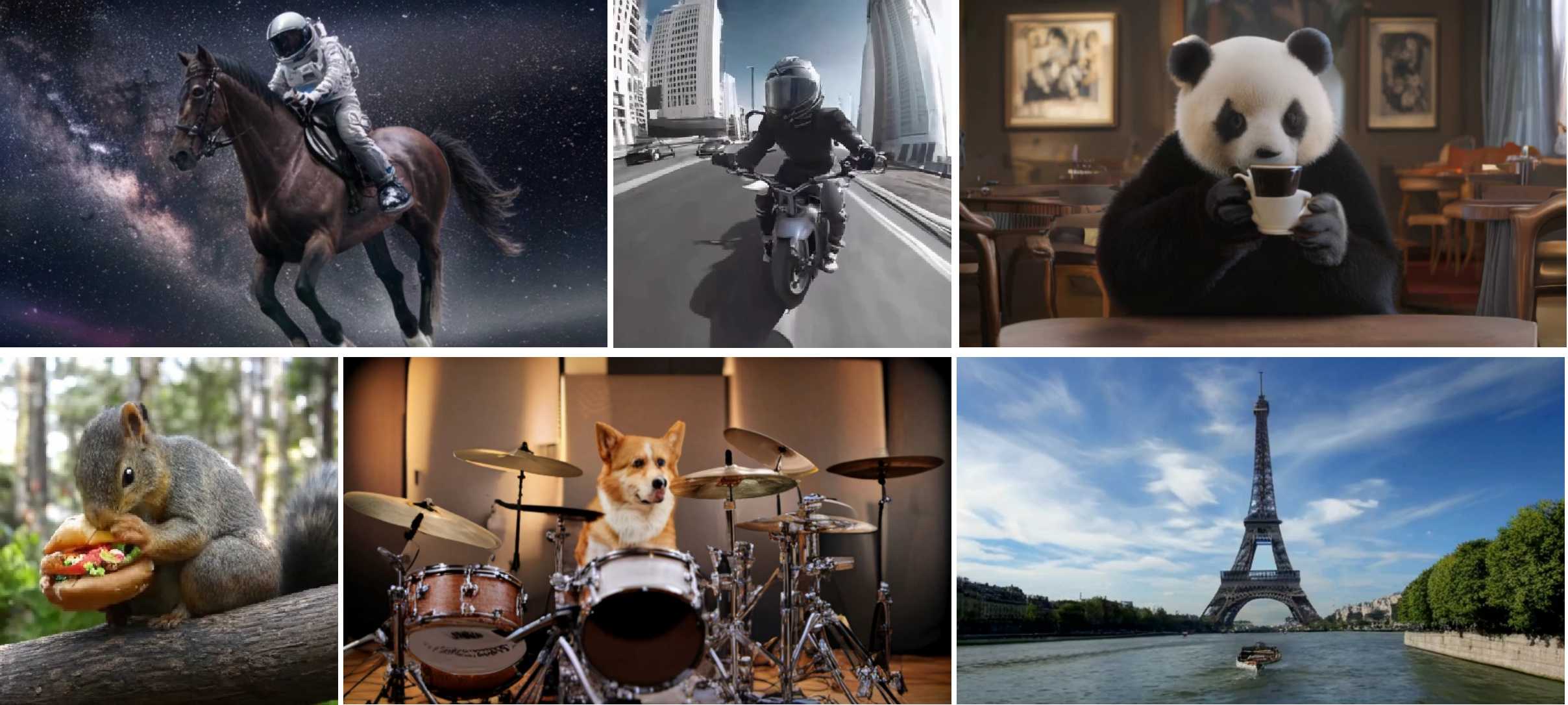}
    \captionsetup{type=figure}
    \vspace{-7mm}
    \captionof{figure}{
    \shortname generates high-fidelity and temporally consistent videos from text prompts.
    }
    \label{fig:teaser}
\end{center}
}]

\input{sec/0_abstract}    
\input{sec/1_intro}

\input{sec/2_relatedwork}

\input{sec/3_method}

\input{sec/4_experiments}

\input{sec/5_conclusion}

\newpage
{
    \small
    \bibliographystyle{ieeenat_fullname}
    \bibliography{main}
}

\input{sec/X_suppl}

\end{document}

%% file: preamble.tex
\usepackage{multirow}
\usepackage{booktabs}
\usepackage{amssymb} 
\usepackage{overpic}
\usepackage{pifont}
\usepackage{makecell}

\usepackage[accsupp]{axessibility}

\newcommand{\shortname}{\emph{VideoAR}\xspace}
\newcommand{\shortnamerope}{\textbf{Multi-scale Temporal RoPE}\xspace}
\newcommand{\shortnamemask}{\textbf{Random Frame Mask}\xspace}
\newcommand{\shortnameflip}{\textbf{Cross-Frame Error Correction}\xspace}
\newcommand{\datasetUCF}{\textit{UCF-101}\xspace}

%% file: sec/0_abstract.tex
\begin{abstract}
Recent advances in video generation have been dominated by diffusion and flow-matching models, which produce high-quality results but remain computationally intensive and difficult to scale. In this work, we introduce \shortname, the first large-scale Visual Autoregressive (VAR) framework for video generation that combines multi-scale next-frame prediction with autoregressive modeling. \shortname disentangles spatial and temporal dependencies by integrating intra-frame VAR modeling with causal next-frame prediction, supported by a 3D multi-scale tokenizer that efficiently encodes spatio-temporal dynamics. To improve long-term consistency, we propose \shortnamerope, \shortnameflip, and \shortnamemask, which collectively mitigate error propagation and stabilize temporal coherence. Our multi-stage pretraining pipeline progressively aligns spatial and temporal learning across increasing resolutions and durations. Empirically, \shortname achieves new state-of-the-art results among autoregressive models, improving FVD on UCF-101 from 99.5 to 88.6 while reducing inference steps by over 10×, and reaching a VBench score of 81.74—competitive with diffusion-based models an order of magnitude larger. These results demonstrate that \shortname narrows the performance gap between autoregressive and diffusion paradigms, offering a scalable, efficient, and temporally consistent foundation for future video generation research.
\end{abstract}

%% file: sec/1_intro.tex
\section{Introduction}
Video generation has achieved remarkable progress in recent years, enabling high-quality synthesis that captures both spatial structures and temporal dynamics. Most commercial and open-source systems are built upon diffusion and flow-matching frameworks~\cite{lipman2022flow}, which typically leverage well pre-trained image diffusion models and are further adapted for temporal consistency. However, large-scale video diffusion models are computationally expensive and difficult to scale, as they rely on bi-directional denoising of entire temporal sequences simultaneously.

In contrast, autoregressive (AR) models have gained traction in image generation due to their scalability, the ease of integrating with optimized LLM infrastructures, and their unified modeling of diverse modalities. A critical step in adapting LLMs for vision is to design effective visual tokenizers that discretize visual inputs. Early pixel-based or VQ-VAE\cite{van2017vqvae}  tokenization approaches adopted the raster-scan next-token prediction paradigm, but they proved inefficient for complex visual signals. More recently, Visual AutoRegressive (VAR) \cite{VAR} modeling reformulated autoregression as a coarse-to-fine “next-scale prediction,” achieving superior generalization and scaling compared to diffusion models while requiring fewer inference steps.

Despite this progress, autoregressive (AR) approaches for video generation remain underexplored  and face several fundamental challenges when applied to high-dimensional spatio-temporal data: 1) \textbf{Mismatch between spatial and temporal modeling} - naive next-token prediction~\cite{wang2024emu3,agarwal2025cosmos} modeling is poorly aligned with the intrinsic structure of video data, as causal temporal progression and 2D spatial synthesis follow different modeling principles~\cite{yu2025videomar}. 2) \textbf{Error propagation} - autoregressive over long video token sequences leads to severe error propagation~\cite{wang2024loong}, resulting in substantial quality degradation. 3) \textbf{Limited temporal–spatial controllability} - existing temporal–spatial sampling strategies lack fine-grained controllability regarding video dynamics and duration.

To address these challenges, we propose \shortname, the first VAR-based framework for large-scale video generation pretraining. \shortname  disentangles spatial and temporal modeling by leveraging the intra-frame modeling strength of VAR while adopting a next-frame prediction paradigm for inter-frame dependencies. For tokenization, we extend the 2D encoder–decoder of VAR into a 3D architecture that captures temporal dynamics, and initialize it with pre-trained 2D-VAR weights to efficiently transfer spatial knowledge. The Transformer backbone autoregressively predicts frames conditioned on text prompts and preceding frames, with each frame represented by multi-scale tokens produced by our 3D-VAR encoder. To better capture spatio-temporal relations, we introduce \shortnamerope, which enhances temporal awareness and improves bit-level prediction accuracy. 
\shortname further adopt two training strategies to address the error propagation problem: (1) \shortnameflip where the flip ratio is progressively increased across frames and inherated between cross-frame transitions, and (2) \shortnamemask, which weakens excessive reliance on previous frames and alleviates over-memorization. Finally, we incorporate temporally-adjustable classifier guidance to flexibly control video dynamics, and employ frame re-encoding to enable duration extension.

\shortname establishes a new state-of-the-art among autoregressive models on video generation tasks.
Notably, \shortname-L achieves a gFVD score of 90.3 on the widely used UCF-101 benchmark, representing a substantial improvement over the previous best autoregressive model, PAR-4×, which reported 99.5 gFVD. This significant reduction in gFVD highlights the strong generative capability of our base model.

Furthermore, we train a large-scale variant on real-world video data, which achieves comparable performance to leading diffusion-based methods such as CogVideo~\cite{hong2022cogvideo} and Step-Video~\cite{Ma2025StepVideoT2VTR} on VBench~\cite{huang2024vbench}.
\shortname can generate 4-second videos at a resolution of 384×672 with high fidelity and strong temporal consistency.
In addition, we observe clear scaling behavior—enlarging the transformer backbone consistently improves the quality of generated videos.

In summary, our main contributions are threefold:
\begin{itemize}
\item We introduce \shortname, the first video generation framework that integrates the VAR paradigm with next-frame prediction, enabling multi-scale and temporally consistent synthesis.

\item We propose \shortnamerope and two effective training strategies—\shortnameflip and \shortnamemask—to enhance spatio-temporal modeling. These techniques are particularly effective in long-form generation, mitigating frame drift and context collapse.

\item Our efficiently pre-trained video tokenizer and transformer backbone achieve state-of-the-art results on standard generation benchmarks, while significantly outperforming prior methods in inference speed and computational efficiency.
\end{itemize}
With these designs, \shortname not only narrows the gap between autoregressive and diffusion-based models, but also establishes a foundation for future large-scale video generation research.

%% file: sec/2_relatedwork.tex
\section{Related Work}

\subsection{Diffusion-based Video Generation}

Recent video generation models such as Veo3~\cite{Veo3}, Sora~\cite{Sora2}, and Wanx~\cite{wan2025} have achieved remarkably realistic visual quality and strong temporal consistency by applying large-scale latent diffusion models~\cite{rombach2022high}. These models progressively synthesize visual content through iterative noise injection and denoising in the latent space.
In parallel, several works~\cite{ai2025magi1autoregressivevideogeneration, deng2025bagel} explore AR-Diffusion, which integrates autoregressive modeling with diffusion processes by introducing a non-decreasing corruption schedule and temporally causal attention, aligning training and inference dynamics for stable image-to-image translation and long-term video generation.
While diffusion-based approaches offer superior fidelity and temporal smoothness, they remain computationally expensive during inference and lack flexibility in controlling generation length or ensuring fine-grained temporal coherence.

\subsection{Autoregressive Visual Generation}
Inspired by the success of AR modeling in natural language processing, recent studies have extended this paradigm to visual domains. 
In contrast to diffusion models, autoregressive (AR) generation predicts visual elements sequentially, making it naturally well-suited for structured sequence modeling tasks such as visual-token and temporal frame-wise generation. LlamaGen~\cite{sun2024autoregressive} and MAGVIT-v2~\cite{yu2023language} adopt a next-token prediction framework over visual tokens obtained from VQ-VAE-style tokenizers, which quantize latent representations into discrete symbols. These methods achieve comparable visual quality to diffusion models while being significantly more efficient at inference time.
PAR~\cite{wang2025parallelized} further parallelizes the next-token prediction process to improve inference efficiency, and Loong~\cite{wang2024loong} pioneers AR-based video generation by introducing temporal-balanced losses and a multi-stage training strategy.
However, token-based AR methods still suffer from low spatial resolution due to excessive token lengths, and their generation quality often degrades because of error accumulation across spatial-temporal dimensions and weak modeling of spatial correlations among flattened pixel tokens.
To address these limitations, we propose \shortname, which combines temporal causal modeling with inter-frame multi-scale spatial generation and cross-frame self-correction, enabling efficient and coherent autoregressive video synthesis.

%% file: sec/3_method.tex
\section{Preliminary}
\label{sec:preliminary}
\textbf{Visual autoregressive (VAR)} model generally consist of a multi-scale visual tokenizer and a Transformer-based generator. An image $\mathbf{I} \in \mathbb{R}^{H \times W \times 3}$ is tokenized with an encoder $\mathcal{E}$ into a feature map $\mathbf{F} \in \mathbb{R}^{H \times W \times d} = \mathcal{E}(\mathbf{I})$,  where $H, W, d$ are the height, width, and channels. A quantizer $\mathcal{Q}$  then decomposes $\mathbf{F}$ into $K$ multi-scale residual maps $\mathbf{R}_{1: K} = \mathcal{Q}(F)$ with gradually increasing resolution, where each residual $\mathbf{R}_k \in \mathbb{R}^{H_k \times W_k \times |V|}$ and $|V|$ is the vocabulary size. The Transformer autoregressively predicts residuals at the next scale conditioned on all previous scales and the text prompt $\Psi$:
\begin{equation}
    p(\mathbf{R}_k|\mathbf{R}_{1:k-1}, \Psi)
\end{equation}
Concretely, to predict $\mathbf{R}_k$, the model take the feature representaion $\mathbf{F}_{k - 1}$ as the input by aggregating all previously generated residuals $\mathbf{R}_{1:k-1}$, upsamples them to the base resolution $(H, W)$, and then downsamples to match the target resolution $(H_k, W_k)$:
\begin{equation}
    \tilde{\mathbf{F}}_{k - 1} = \text{down}(\sum_{i=1}^{k-1} \text{up}(\mathbf{R}_i, (H, W)), (H_{k}, W_{k}))
\end{equation}
where \textit{up} and \textit{down} denotes bilinear up- and down-sampling respectively. 

During training, to mitigate train–test discrepancies and improve robustness to prediction errors, the transformer takes as input the partially corrupted version of $\mathbf{F}_k^{\text{flip}}$. Specifically, with probability $p_{\text{flip}}$, some labels in $\mathbf{R}_k$ are randomly flipped:
\begin{equation}
    \mathbf{R}_k^{\text{flip}} = p_{\text{flip}}(\mathbf{R}_k)
\end{equation}
This enables the model to account for potential prediction errors and better generalize to autoregressive inference.

\section{Method}
\begin{figure*}[!ht]
\centering
    \begin{overpic}[width=1.0\textwidth]{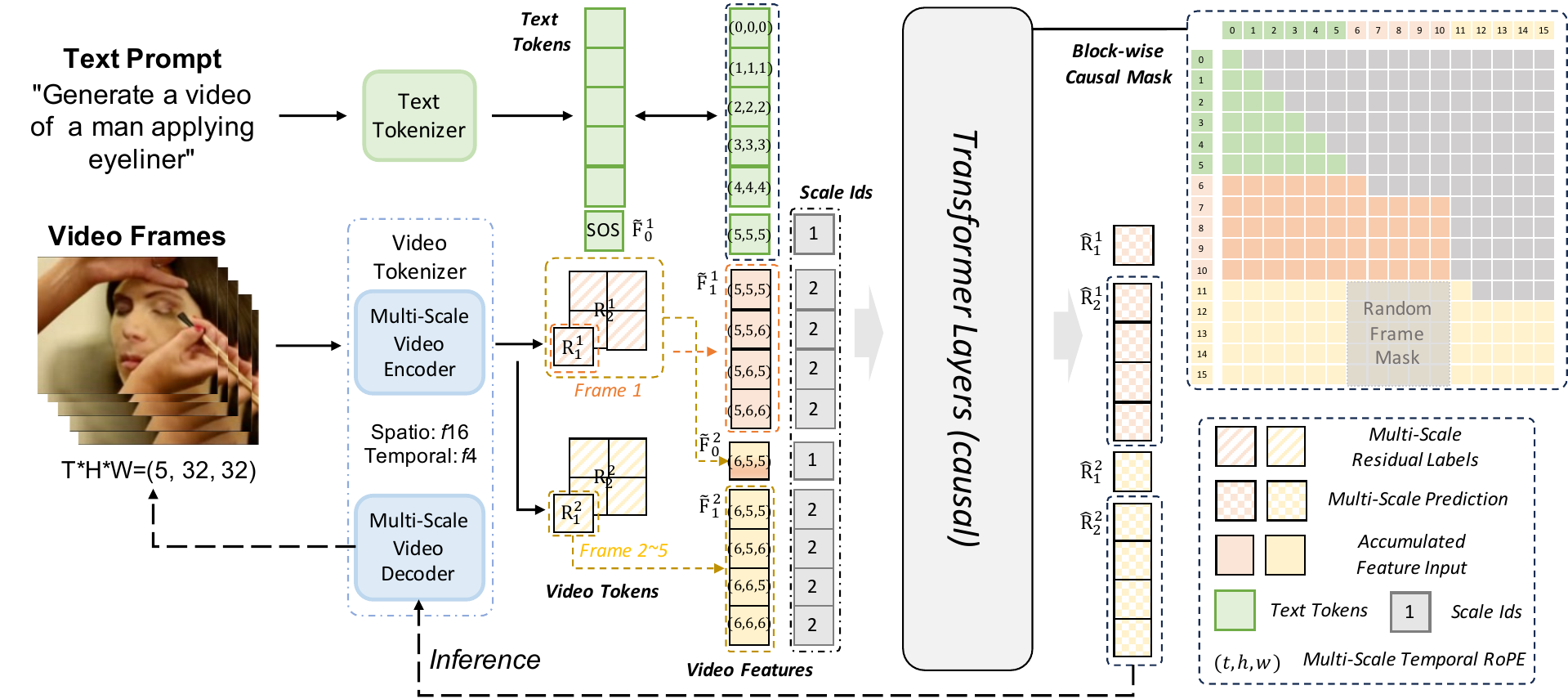}
    \end{overpic}
    \vspace{-2mm}
    \caption{Overall framework of \shortname. 
Given a text prompt, the video frames are first compressed into a sequence of spatio-temporal tokens via a \textbf{multi-scale causal 3D tokenizer}. 
Each frame is represented by residual maps at multiple scales, which are autoregressively predicted by a Transformer with block-wise causal masking. 
The input embeddings combine text tokens, accumulated video features, and scale embeddings, while the proposed \textit{Multi-Scale Temporal RoPE} encodes temporal, spatial, and scale-aware positional information. 
Random frame masking is applied during training to mitigate exposure bias and improve long-term consistency. 
Finally, the multi-scale video decoder reconstructs the video frames from the predicted residuals.
}
    \label{fig:framework}
    \vspace{-4mm}
\end{figure*}

In this section, we present our proposed framework \shortname, which integrates the strengths of visual autoregressive (VAR) modeling with next-frame prediction for efficient and high-quality video generation. The pipeline is composed of two main components. First, we introduce a 3D video tokenizer in \cref{subsec:video-tknz} that compresses raw video into compact discrete representations while preserving both spatial and temporal structures. This tokenizer serves as the foundation for scalable and efficient modeling. Second, we design an autoregressive video model in \cref{subsec:ar-video-modeling} built upon multi-scale residual prediction, where temporal consistency is further enhanced by our proposed training strategies. 
\label{sec:method}
\subsection{Visual Tokenizer}\label{subsec:video-tknz}
\noindent\textbf{3D Architecture}. To better capture spatial–temporal correlations, we adopt a \textit{causal 3D convolutional architecture}~\cite{yu2023magvit}, which allows the tokenizer to process both images and videos within a unified framework. Concretely, the 3D convolutional encoder with temporal subsampling compresses the input video $\mathbf{V} \in \mathbb{R}^{ (1 + T) \times H \times W \times 3}$ into a compact spatio-temporal latent representation $\mathbf{F} \in \mathbb{R}^{(1 + T/\tau) \times H' \times W' \times d}$, where $\tau$ denotes the temporal compression factor. This design leverages the inherent redundancy across adjacent frames, enabling efficient video modeling while maintaining fidelity.

To further scale to long-form video generation, we eliminate all non-causal temporal operations (e.g., temporal normalization) from both the encoder and decoder, ensuring that each latent feature only depends on past frames. This causal design enables inference on extremely long videos in a chunk-by-chunk manner, without performance loss compared to full-sequence inference.

\noindent\textbf{Quanization} Considering our temporal-causal modeling, we leverage temporal-independent quantization where each frame is passed through isolated the multi-scale quantizer.    

\noindent\textbf{Training}. To enable efficient and stable training of our video tokenizer, we adopt a \textit{3D inflation strategy} by initializing the model from a well-trained image tokenizer~\cite{han2025infinity}. This initialization provides a strong spatial prior, substantially stabilizing optimization and accelerating convergence. Concretely, following the inflation procedure in~\cite{yu2023language}, we populate the temporally last slice of the 3D CNN using the weights from the image tokenizer, while the remaining temporal parameters and the discriminator are randomly initialized.

The tokenizer is trained with a standard combination of complementary objectives. We apply the \textit{reconstruction}, \textit{perceptual}, \textit{commitment losses} on each frame. Following ~\cite{yu2023magvit}, we use LeCAM regularization~\cite{tseng2021regularizing} for improved stability and entropy penalty to encourage codebook utilization.

The overall training objective can be formulated as:
\begin{equation}
\begin{aligned}
\mathcal{L} &= \lambda_{\text{rec}} \mathcal{L}_{\text{rec}} +
\lambda_{\text{perc}} \mathcal{L}_{\text{perc}} +
\lambda_{\text{GAN}} \mathcal{L}_{\text{GAN}} \\
&\quad +
\lambda_{\text{commit}} \mathcal{L}_{\text{commit}} +
\lambda_{\text{entropy}} \mathcal{L}_{\text{entropy}},
\end{aligned}
\end{equation}
where $\lambda$’s are balancing weights for different objectives.

This training scheme ensures that the tokenizer learns compact yet expressive spatio-temporal representations, benefiting both reconstruction fidelity and downstream autoregressive video generation.

\subsection{Autoregressive Video Modeling}
\label{subsec:ar-video-modeling}

\noindent\textbf{Extension to 3D Architecture}.
Building upon the spatio-temporal features extracted by our 3D tokenizer, we extend the visual autoregressive (VAR) paradigm from images to videos. Specifically, the Transformer autoregressively predicts the residuals of the $t$-th frame conditioned on all previously generated frames, the coarser scales of the current frame, and the text prompt:
\begin{equation}
    p(\mathbf{R}^t_k|\mathbf{R}^{1:t-1}_{1:K}, \mathbf{R}^{t}_{1:k-1}, \Psi)
\end{equation}
where $\mathbf{R}^{1:t-1}_{1:K}$ denotes the multi-scale residual maps of all past frames, and $\mathbf{R}^{t}_{1:k-1}$ denotes the residuals of the already generated coarser scales of the $t$-th frame. The input feature for the $t$-th frame at scale $k$ is constructed as:
\begin{equation}
    \tilde{\mathbf{F}}^t_{k - 1} = \text{down}(\sum_{i=1}^{k-1} \text{up}(\mathbf{R}^t_i, (H, W)), (H_{k}, W_{k}))
\end{equation}
where $\text{up}(\cdot)$ and $\text{down}(\cdot)$ denote spatial up- and down-sampling.  

To initialize generation, the feature of the first scale in the first frame ($\tilde{\mathbf{F}}^1_{0}$ in Fig.~\ref{fig:framework}) is set to a special $<\text{SOS}>$ token embedding, enabling text-conditioned generation. For subsequent frames ($t>1$), the first-scale feature ($\tilde{\mathbf{F}}^t_{0}$) is initialized from the accumulated features of the previous frame, injecting temporal context into the next-frame generation.  
\\

\noindent\shortnamerope.
To better capture spatio-temporal dependencies, we introduce \shortnamerope, an extension of Rotary Position Embeddings (RoPE)~\cite{su2024roformer} by factorizing the embedding space into three axes—time, height, and width. The design principles of \shortnamerope are threefold: (1) compatibility with the native RoPE formulation for text tokens, (2) explicit temporal awareness, and (3) spatial consistency across frames with multi-scale inputs.  

Given a multimodal input consisting of a text prompt $\Psi$ and video tokens, we assign the same temporal, height, and width indices to text tokens to maintain compatibility with RoPE. Let $x_k^t(h, w)$ denote the token at scale $k$ of the $t$-th frame at spatial location $(h,w)$, where $h < H_k$ and $w < W_k$. The positional encoding is defined as:
\begin{equation}
\begin{split}
    \text{Position: }(t, h, w) &= (t + |\Psi|,\; h + |\Psi|,\; w + |\Psi|), \\
    \text{Embedding: } \mathbf{x}_k^t(h, w) &= \tilde{\mathbf{F}}^t_{k}(h, w) + \mathbf{s}_{k},
\end{split}
\end{equation}
where the spatial indices $(h,w)$ are consistent across frames, while the temporal index increases with $t$ to preserve ordering. Additionally, a learnable scale embedding $\mathbf{s}_k$ is added to differentiate coarse-to-fine scales during autoregressive generation.
\\

\noindent\textbf{Temporal-Consistency Enhancement}. 
Autoregressive video generation suffers from error accumulation: quality degrades as $t$ grows due to train--test discrepancy. We adopt two complementary strategies to mitigate this: \emph{\shortnameflip} with a time-ramped schedule, and \emph{\shortnamemask} with a causal sliding window.

\begin{figure}[!ht]
\centering
    \begin{overpic}[width=0.45\textwidth]{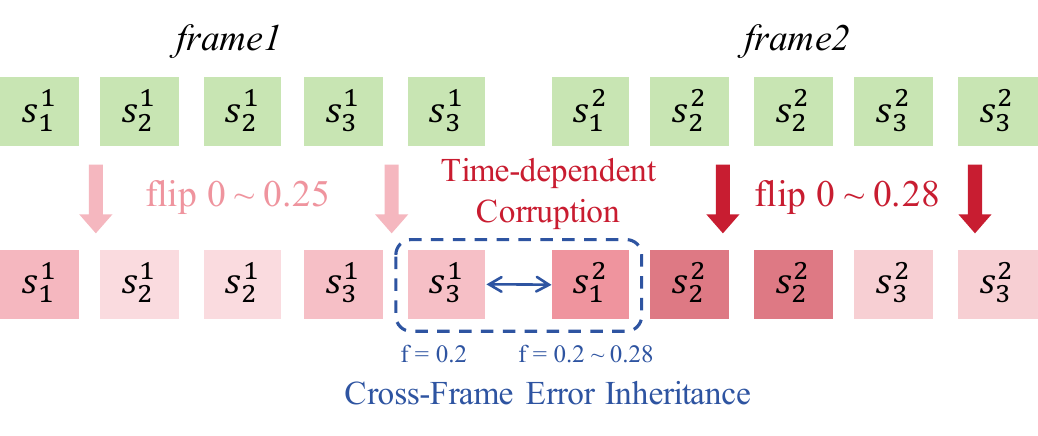}
    \end{overpic}
    \vspace{-2mm}
    \caption{Our proposed Cross-Frame Error Correction. 
}
    \label{fig:cross-frame}
    \vspace{-4mm}
\end{figure}

\noindent\shortnameflip. Following the bitwise formulation in Infinity, we represent each token in $\mathbf{R}^t_k$ by $d$ bits
$b^{t,k}_{h,w,j} \in \{0,1\}$, $j=1,\dots,d$.

To account for the accumulation of error propagation along extended frame sequences, we introduce \textit{time-dependent corruption} by injecting perturbations with progressively increasing flip ratios, thereby simulating inference-time situation, see~\cref{fig:cross-frame}. Furthermore, since errors at the final scale of each frame inevitably propagate into the first scale of the subsequent frame, we propose a \textit{cross-frame error inheritance mechanism}. Specifically, the flip ratio of each frame’s first scale is initialized within a range above the final scale’s flip ratio of the preceding frame. By compelling the model to correct these inherited perturbations at the very first scale, our training procedure enhances temporal robustness and substantially mitigates the influence of preceding-frame errors on subsequent generations.

\begin{equation}
\label{eq:time-flip}
\begin{split}
p_{\text{flip}}(t) &\sim \text{Uniform}(p_{min} + \delta t, p_{max} + \delta t) \\
p_{\text{flip}}(t_{init-f}) &\sim \text{Uniform}(p_{prev-f}, p_{max} + \delta t) \\
\tilde{b}^{t,k}_{h,w,j} &= b^{t,k}_{h,w,j} \oplus \xi^{t,k}_{h,w,j},\ \ \xi^{t,k}_{h,w,j} \sim \mathrm{Bernoulli}\!\bigl(p_{\text{flip}}(t)\bigr),
\end{split}
\end{equation}s
where $\oplus$ denotes XOR, $\delta$ denotes the factor for increasing the flipping range. The model conditions on corrupted history and is supervised by self-corrected targets with re-quantized errors~\cite{han2025infinity}, improving robustness to compounding mistakes.

\noindent\shortnamemask
Let the attention window size be $w$. For each step $t$, we form a stochastic causal context
$S_t = \{\,t'\,|\, t' \in [\max(1,t-w),\,t-1],\ m_{t'}=1\,\}$ with i.i.d.
$m_{t'} \sim \mathrm{Bernoulli}(1-p_{\text{mask}})$.
Denote text keys/values by $(\mathbf{K}_{\text{text}}, \mathbf{V}_{\text{text}})$
and video keys/values from frames in $S_t$ by
$(\mathbf{K}^{S_t}, \mathbf{V}^{S_t})$.
The attention output for frame $t$ is
\begin{equation}
\label{eq:rand-mask-attn}
\mathbf{O}^t
= \mathrm{Softmax}\!\left(
\frac{\mathbf{Q}^t\,[\mathbf{K}_{\text{text}},\, \mathbf{K}^{S_t}]^\top}{\sqrt{d}}
\right)
[\mathbf{V}_{\text{text}},\, \mathbf{V}^{S_t}],
\end{equation}
which discourages over-reliance on distant frames while preserving necessary temporal context.\\

\noindent\textbf{Multi-Stage Training Pipeline.} 
Following Infinity~\cite{han2025infinity}, our training objective is defined as a bit-wise cross-entropy loss between the predicted residual maps $\hat{\mathbf{R}}_{1:K}^{1:T}$ and the ground truth $\mathbf{R}_{1:K}^{1:T}$. 
To achieve robust temporal consistency and high-quality synthesis in long-form, high-resolution videos, we adopt a progressive multi-stage training strategy. 
In \textit{Stage I}, we jointly pretrain on large-scale image and low-resolution video datasets, enabling the model to acquire fundamental spatial-temporal representations while benefiting from efficient convergence. 
In \textit{Stage II}, we continue training on higher-resolution image and video data to enhance fine-grained visual fidelity and temporal coherence. 
Finally, in \textit{Stage III}, we perform long-form video fine-tuning using only high-resolution video datasets, allowing the model to capture extended motion dynamics and long-range temporal dependencies. 
This hierarchical training scheme effectively balances training stability, scalability, and generation quality across diverse video domains.\\

\begin{figure*}[!ht]
\centering
    \begin{overpic}[width=1\textwidth]{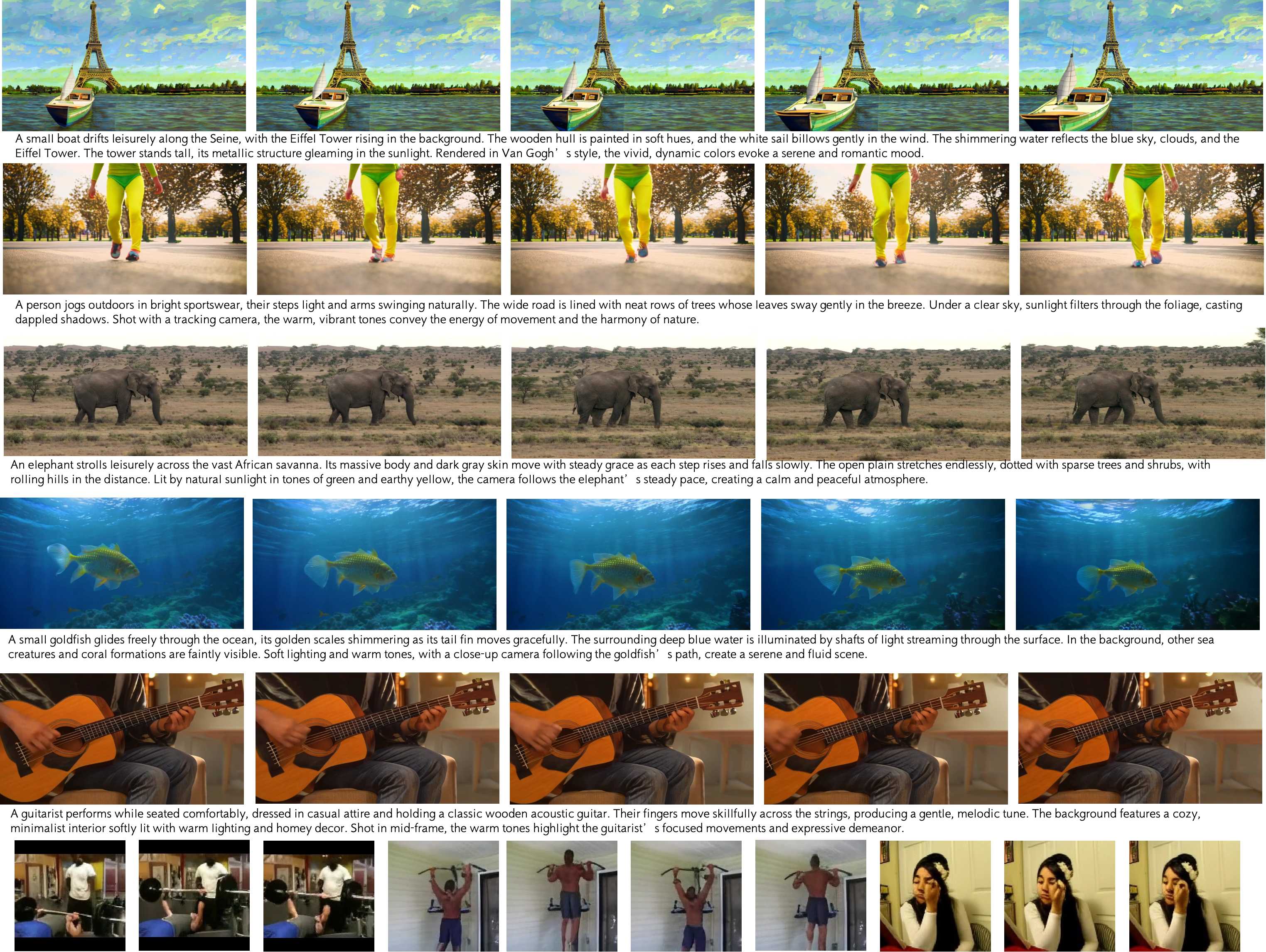}
    \end{overpic}
    \vspace{-4mm}
    \caption{Generation results from our \shortname-4B on VBench and \datasetUCF datasets. Zoom in for clearer visualization.
}
    \label{fig:demo}
    \vspace{-4mm}
\end{figure*}

\noindent\textbf{Temporal-Spatial Adaptive Classifier-free Guidance}.
At test time, we perform causal decoding over $(t,k)$ with cached states to ensure efficiency. To balance semantic fidelity and temporal consistency, we introduce a temporal-spatial adaptive classifier-free guidance (CFG) applied to the logits to enable flexible control of text alignment and temporal dynamics under different model settings.

Empirically, we observe that larger guidance coefficients lead to improved visual quality and stronger dynamics across frames, whereas smaller coefficients yield more stable temporal transitions and greater sampling diversity. Therefore, we adapts spatial-CFG along scales and also set the temporal starting point of CFG for first scale in a pre-selected scheduler.

%% file: sec/4_experiments.tex
\begin{figure*}[t]
\centering
{
\includegraphics[width=1.9\columnwidth]{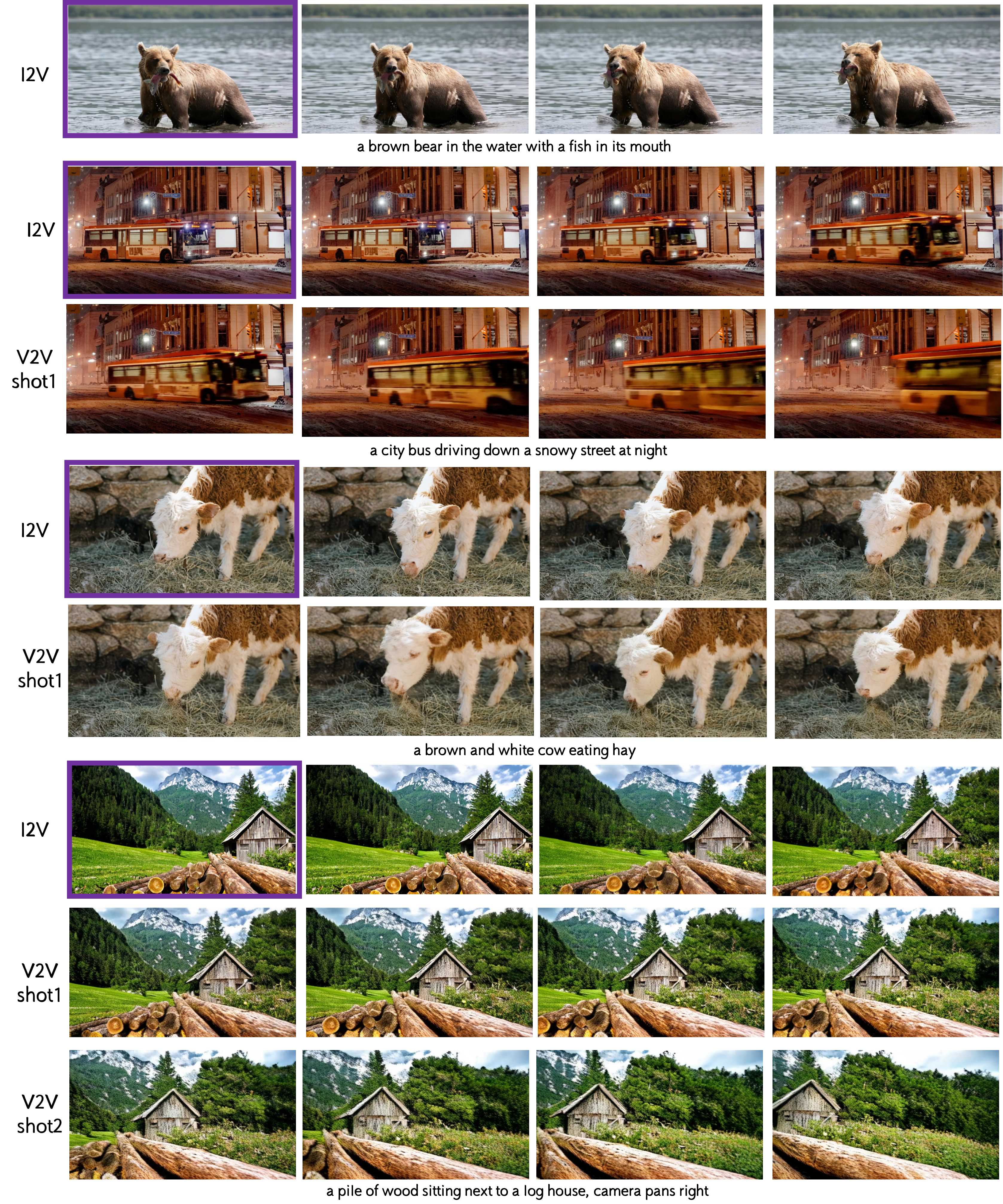} }
\vspace{-3mm}
\caption{
Visualization for VideoAR's Image-to-Video and Video-to-Video generation performance. I2V refers to Image-to-Video, purple boxes refers to the given image. V2V shot$N$ refers to N times video-to-video Extension of 4 seconds window.}
\label{suppl_show_i2v}
\vspace{-0.2cm}
\end{figure*}

\section{Experiments}
\subsection{Experimental Setup}  

\noindent\textbf{Datasets.} 
We conduct experiments on a diverse set of benchmarks encompassing both low-resolution toy datasets and high-resolution real-world long-form video generation.
For short videos generation, we use \datasetUCF~\cite{soomro2012ucf101} (8K clips, 101 action categories) as a standard benchmark for human-action modeling. 
For long-form and open-domain scenarios, we conduct large-scale pretraining and evaluation on proprietary in-house datasets.
All videos are uniformly resized to $pixel_{sqrt} \in (128, 256, 512) $ and temporally sampled to $T \in [17, 65]$ frames depending on the dataset.

\noindent\textbf{Evaluation metrics.} 
We evaluate our model along two axes: \emph{reconstruction quality} and \emph{generation quality}. 
For reconstruction, we report the Fréchet Video Distance (rFVD)~\cite{unterthiner2019fvd}, which directly reflects the fidelity of the learned video tokenizer. 
For generation quality, we measure gFVD on the held-out human-centric test set of \datasetUCF. Moreover, to assess real-world generation performance, we evaluate on the standard \textit{VBench}~\cite{huang2024vbench}, which provides a comprehensive suite of perceptual and temporal metrics specifically designed for video generation models.

\begin{table*}[!t]
\centering
\caption{
    \textbf{Performance comparison on VBench.} The best results are in \textbf{bold}, and the second-best are \underline{underlined}. Our 4B model achieves a competitive overall score and sets a new state-of-the-art on the Semantic Score, Aesthetic Qualiy, Object Class and Multiple Objects, surpassing models with substantially larger parameters. 
}
\vspace{-2mm}
\renewcommand{\arraystretch}{1.1}
\sisetup{detect-weight=true, detect-family=true} 
\resizebox{\linewidth}{!}{
\begin{tabular}{l c c c c c c c c c c c c c c c c c c c c}
\toprule

\multicolumn{1}{l}{\textbf{Methods}} & {\textbf{\#Size}} & {\textbf{Total}} &
\makecell{\scriptsize Quality\\\scriptsize Score} &
\makecell{\scriptsize Semantic\\\scriptsize Score} &
\makecell{\scriptsize Subject\\\scriptsize Consis} &
\makecell{\scriptsize Background\\\scriptsize Consis} &
\makecell{\scriptsize Temp\\\scriptsize Flicker} &
\makecell{\scriptsize Motion\\\scriptsize Smooth} &
\makecell{\scriptsize Dynamic\\\scriptsize Degree} &
\makecell{\scriptsize Aesthetic\\\scriptsize Quality} &
\makecell{\scriptsize Image\\\scriptsize Quality} &
\makecell{\scriptsize Object\\\scriptsize Class} &
\makecell{\scriptsize Multiple\\\scriptsize Objects} &
\makecell{\scriptsize Human\\\scriptsize Action} &
\makecell{\scriptsize Color} &
\makecell{\scriptsize Spatial\\\scriptsize Relation} &
\makecell{\scriptsize Scene} &
\makecell{\scriptsize Appearance\\\scriptsize Style} &
\makecell{\scriptsize Temporal\\\scriptsize Style} &
\makecell{\scriptsize Overall\\\scriptsize Consis} \\

\midrule
LaVie\cite{wang2023lavie}          & 3B  & 77.08 & 78.78 & 70.31 & 91.41 & 97.47 & 98.30 & 96.38 & 49.72 & 54.94 & 61.90 & 91.82 & 33.32 & \underline{96.80} & 86.39 & 34.09 & 52.69 & 23.56 & \underline{25.93} & 26.41 \\
VideoCrafter-2.0\cite{chen2024videocrafter2} & 1B  & 80.44 & 82.20 & 73.42 & 96.85 & \textbf{98.22} & 98.41 & 97.73 & 42.50 & 63.13 & 67.22 & \underline{92.55} & 40.66 & 95.00 & \underline{92.92} & 35.86 & \textbf{55.29} & \textbf{25.13} & 25.84 & \textbf{28.23} \\
CogVideoX\cite{hong2022cogvideo}       & 5B  & 81.61 & 82.75 & \underline{77.04} & 96.23 & 96.52 & 98.66 & 96.92 & \underline{70.97} & 61.98 & 62.90 & 85.23 & 62.11 & \textbf{99.40} & 82.81 & 66.35 & 53.20 & \underline{24.91} & 25.38 & \underline{27.59} \\
Kling\cite{kling}           & -   & 81.85 & 83.39 & 75.68 & \textbf{98.33} & 97.60 & \underline{99.30} & \textbf{99.40} & 46.94 & 61.21 & 65.62 & 87.24 & 68.05 & 93.40 & 89.90 & \textbf{73.03} & 50.86 & 19.62 & 24.17 & 26.42 \\
Step-Video-T2V\cite{Ma2025StepVideoT2VTR}  & 30B & 81.83 & \underline{84.46} & 71.28 & 98.05 & 97.67 & \textbf{99.44} & 99.08 & 53.06 & 61.23 & \textbf{70.63} & 80.56 & 50.55 & 94.00 & 88.25 & \underline{71.47} & 24.38 & 23.17 & \textbf{26.01} & 27.12 \\
Gen-3\cite{gen3}           & -   & \underline{82.32} & 84.11 & 75.17 & 97.10 & 96.62 & 98.61 & \underline{99.23} & 60.14 & \underline{63.34} & 66.82 & 87.81 & 53.64 & 96.40 & 80.90 & 65.09 & \underline{54.57} & 24.31 & 24.71 & 26.69 \\
Hunyuan-Video\cite{kong2024hunyuanvideo}   & 13B & \textbf{83.24} & \textbf{85.09} & 75.82 & 97.37 & 97.76 & \textbf{99.44} & 98.99 & \textbf{70.83} & 60.36 & \underline{67.56} & 86.10 & \underline{68.55} & 94.40 & \underline{91.60} & 68.68 & 53.88 & 19.80 & 23.89 & 26.44 \\
\midrule
\textbf{\shortname (Ours)} & 4B  & 81.74 & 82.88 & \textbf{77.15} & 95.51 & \underline{97.85} & 98.83 & 98.37 & 61.39 & \textbf{63.42} & 60.71 & \textbf{94.98} & \textbf{72.88} & 94.40 & 90.73 & 56.14 & 50.52 & 22.68 & 25.11 & 27.44 \\
\bottomrule
\end{tabular}
}
\label{tab:vbench_results}
\end{table*}

\subsection{Experimental results}

\begin{table}[!t]
\centering
\caption{Reconstruction performance of our video tokenizer and other methods on the \datasetUCF dataset.}
\vspace{-2mm}
\renewcommand{\arraystretch}{1.1}
\resizebox{0.9\linewidth}{!}{
\begin{tabular}{lccc}
\hline
\multicolumn{1}{l}{\textbf{Methods}} & \textbf{Tokens (T$\times$H$\times$W)} & \textbf{Ratio} & \textbf{rFVD $\downarrow$} \\ \hline
TATS~\cite{ge2022long}   & $4 \times 16 \times 16$  & 8     & 162       \\ 
MAGVIT~\cite{yu2023magvit}       & $5 \times 16 \times 16$  & 8     & 58        \\ 
OmniTokenizer~\cite{wang2024omnitokenizer} & $5 \times 16 \times 16$  & 8     & 42        \\ \hline
\textbf{\shortname-L (Ours)} & \textbf{$5 \times 8 \times 8$}    & \textbf{16}    & \textbf{61}        \\ \hline
\end{tabular}
}
\label{tab:tokenizer_performance}
\end{table}

\noindent\textbf{Video Reconstruction.}
The efficacy of an autoregressive video generation model is largely depends on the quality and compactness of its underlying video tokenizer. We assess this aspect by reporting the reconstruction Fréchet Video Distance (rFVD). \cref{tab:tokenizer_performance} presents a comparative analysis on the \datasetUCF dataset, demonstrating our model's superior trade-off between compression efficiency and reconstruction fidelity.

Our \shortname-L tokenizer employs an aggressive 16$\times$ spatial compression, encoding video clips into a compact $5 \times 8 \times 8$ latent token grid. This design yields a \textbf{4$\times$ reduction in sequence length} compared to recent state-of-the-art video tokenizers such as MAGVIT~\cite{yu2023magvit} and OmniTokenizer
\cite{wang2024omnitokenizer}, both of which operate at only 8$\times$ compression ratio. Despite the substantially lower token density, our tokenizer maintains excellent reconstruction quality, achieving an rFVD score of 61—on par with MAGVIT (58). This result highlights the effectiveness of our tokenizer in retaining fine-grained spatial and temporal structure, establishing a strong and efficient representation for downstream autoregressive video generation.

\begin{table}[!t]
\centering
\caption{Video generation results: class-conditional generation on \datasetUCF dataset. \shortname achieves the best gFVD score with significantly fewer steps and lower latency.}
\vspace{-2mm}
\renewcommand{\arraystretch}{1.1}
\resizebox{\linewidth}{!}{
\begin{tabular}{lcccc}
\hline
\multicolumn{1}{l}{\textbf{Methods}} & \textbf{\#Params} & \textbf{gFVD $\downarrow$} & \textbf{Steps $\downarrow$} & \textbf{Time (s) $\downarrow$} \\ \hline
CogVideo~\cite{hong2022cogvideo}        & 9.4B         & 626          & -            & -           \\ 
TATS~\cite{ge2022long}            & 312M         & 332          & -            & -           \\ 
OmniTokenizer~\cite{wang2024omnitokenizer}   & 650M         & 191          & 5120         & 336.70      \\ 
MAGVIT-v2-AR~\cite{yu2023language}    & 840M         & 109          & 1280         & -           \\ 
PAR-4x~\cite{wang2025parallelized}          & 792M         & 99.5         & 323          & 11.27       \\ \hline
\textbf{\shortname-L (Ours)} & 926M         & \textbf{90.3} & \textbf{30}  & \textbf{0.86} \\ 
\textbf{\shortname-XL (Ours)}& 2.0B         & \textbf{88.6} & \textbf{30}  & \textbf{1.12} \\ \hline
\end{tabular}
}
\vspace{-0.3cm}
\label{tab:ucf_sota}
\end{table}

\noindent\textbf{Video Generation on \datasetUCF.}
Our \shortname framework establishes a new state-of-the-art on the \datasetUCF dataset, marking a paradigm shift in achieving both superior generation quality and unprecedented inference efficiency.
As shown in \cref{tab:ucf_sota}, our 2B parameter model, \shortname-XL, achieves a new best FVD of \textbf{88.6}, surpassing the previous leading autoregressive model, PAR-4x, by a notable 11\%. Even our smaller 926M model, \shortname-L, outperforms it with an FVD of \textbf{90.3}. The most remarkable advancement, however, lies in inference speed: with only \textbf{30} decoding steps—a more than 10$\times$ reduction—\shortname-L generates a video in just \textbf{0.86 seconds}, achieving over 13$\times$ faster inference compared to PAR-4x.

This dual advancement stems directly from our architectural innovations. High-fidelity spatial details are preserved through intra-frame visual autoregression, while robust temporal consistency.\\

\noindent\textbf{Real-World Video Generation.} To further validate the effectiveness and scalability of our approach, we pre-train a 4B-parameter \shortname model on the challenging task of real-world video generation. As presented in \cref{tab:vbench_results}, our model attains an overall VBench score of 81.74, achieving performance comparable to, or even surpassing, current state-of-the-art models that are significantly larger in scale, such as Step-Video-T2V (30B) and Hunyuan-Video (13B).

Our model’s primary strengths are revealed through a fine-grained analysis of the VBench metrics. In particular, \shortname achieves a new state-of-the-art Semantic Score (SS) of 77.15, surpassing all competitors. This result highlights its remarkable ability to maintain precise text-to-video alignment. While maintaining competitive results on general visual quality metrics such as Aesthetic Quality (AQ) and Overall Consistency (OC), these superior performances in semantics and motion clearly showcase the distinctive strengths of our model.

Qualitative results (\cref{fig:demo} and supplementary material) further corroborate the quantitative improvements. \shortname consistently produces visually compelling and semantically coherent videos, spanning imaginative artistic stylizations, high-fidelity natural scenes, and dynamic human actions with strong temporal consistency.

Crucially, these results confirm that our \shortname strategy offers a compelling alternative to diffusion-based paradigms.  It achieves SOTA-level performance, particularly in semantic control and motion depiction, while providing strong potential for improved scalability and significantly higher inference efficiency.

\noindent\textbf{Image-to-Video and Video-Continuation}.

As an autoregressive video generation model, our proposed VideoAR can directly extend future frames from preceding content (including an initial image and sequential frames) without requiring external fine-tuning. For evaluation, we sample several test cases from VBench-I2V. We present multiple Image-to-Video (I2V) and Video-to-Video (V2V) examples where VideoAR enables single or multi-shot continuous video generation.
As shown in Figure~\ref{suppl_show_i2v}, VideoAR-4B accurately follows semantic prompts aligned with input images across various settings, including object motion control and camera trajectory adjustments. For the video continuity task, VideoAR can generate natural and consistent content over multiple iterations, ultimately producing long-form videos exceeding 20 seconds in duration.

\begin{table}[!t]
\centering
\caption{Ablation study for \shortnamerope and \shortnameflip on the \datasetUCF dataset. The checkmark (\ding{51}) indicates the component is enabled.}
\vspace{-2mm}
\label{tab:ablation}
\renewcommand{\arraystretch}{1.2}
\resizebox{1\linewidth}{!}{
\begin{tabular}{lcccc}
\toprule
\multicolumn{1}{l}{\textbf{Methods}} & \textbf{\shortstack[c]{Multi-scale Temp. \\ RoPE}} & \textbf{\shortstack[c]{Time-Dependent \\ Corruption}} & \textbf{\shortstack[c]{Error \\ Inheritance}} & \textbf{gFVD} $\downarrow$ \\
\midrule
\shortname-L (Baseline) & & & & 96.04 \\
\shortname-L & \ding{51} & & & 94.95 \\
\shortname-L & \ding{51} & \ding{51} & & 93.57 \\
\shortname-L (Full) & \ding{51} & \ding{51} & \ding{51} & \textbf{92.50} \\
\bottomrule
\end{tabular}
}
\vspace{-0.6cm}
\end{table}

\subsection{Ablation Studies}
We conduct a comprehensive ablation study on the \datasetUCF dataset. All models are trained for a fixed 1,000 steps, which is sufficient to reveal clear trends in model performance.

\noindent\textbf{Effect of \shortnamerope}. Our first enhancement replaces the standard positional encoding with \shortnamerope. As shown in the second row of \cref{tab:ablation}, this single modification reduces the FVD from 96.04 to 94.95. This result highlights the importance of rotational relative positional encoding for modeling the complex spatio-temporal dynamics of video data, thereby improving frame-to-frame consistency.

\noindent\textbf{Effect of Temporal-Consistency Enhancement}.Next, we evaluate our proposed \shortnameflip mechanism, which consists of two synergistic components.
(1) We first activate Time-dependent Corruption, a data augmentation strategy that simulates inference-time conditions during training. This addition further reduces the FVD to 93.57.
(2) Building on this, we incorporate Error Inheritance Initialization, which encourages the model to correct inherited perturbations for improved future predictions. This final step yields our full model, achieving a state-of-the-art FVD of \textbf{92.50}.
Further ablation for \shortnamemask is performed on our large-scale real-world dataset, as strong augmentation to small datasets \datasetUCF can impede model convergence. As shown in ~\cref{tab:ablation_mask}, incorporating this technique during the 256px training stage improves the overall VBench score from 76.22 to \textbf{77.00}.

\begin{table}[!t]
\centering
\caption{Ablation study for \shortnamemask on the VBench benchmark during 256px stage-I training.}
\vspace{-2mm}
\label{tab:ablation_mask}
\renewcommand{\arraystretch}{1.2}
\resizebox{1\linewidth}{!}{
\begin{tabular}{lccc}
\toprule
\multicolumn{1}{l}{\textbf{Methods}} & \textbf{Overall Score} & \textbf{Quality} & \textbf{Semantic} \\
\midrule
\shortname-L  & 76.22 & 78.63 & \textbf{66.64} \\
\shortname-L w/ Rand. Mask  & \textbf{77.00} & \textbf{79.78} & 65.89 \\
\bottomrule
\end{tabular}
}
\vspace{-0.4cm}
\end{table}

%% file: sec/5_conclusion.tex
\section{Discussion}
\subsection{Comparsion to Concurrent work: InfinityStar}
We highlight several key differences compared to InfinityStar \cite{liu2025infinitystar}.

\textbf{(1) Spatio-temporal Modeling Paradigm.}  
InfinityStar adopts a 3D-VAR formulation, where each generation block operates over a temporal window of frames. In contrast, our VideoAR employs a \emph{next-frame prediction} paradigm combined with \emph{multi-scale modeling within each frame}. This design enables fine-grained spatial modeling through structured coarse-to-fine generation, while maintaining temporal consistency via explicit frame-wise prediction. 

\textbf{(2) Training Strategy.}  
InfinityStar is fine-tuned from a well-established 8B-scale image generation foundation model, benefiting from strong pre-trained priors. In contrast, our VideoAR is trained from scratch using joint low-resolution image–video data, which focuses on learning unified spatio-temporal representations from the ground up. 

\textbf{(2) Training Scale and Sequence Length.} 
Also, VideoAR is trained with relatively modest sequence lengths, primarily due to practical training considerations at this stage. Consequently, long-horizon temporal coherence has not yet been exhaustively explored. Nevertheless, the proposed framework imposes no inherent limitation on sequence length, and is fully compatible with longer-context training. We expect further gains in long-term consistency as training scale and sequence length are increased.

\section{Conclusion}
We present \shortname, a new paradigm for scalable autoregressive video generation built upon the next-scale prediction principle. By extending VAR framework to videos, \shortname unifies spatial and temporal modeling through a causal 3D tokenizer and a Transformer-based generator. The proposed \shortnamerope enhances spatio-temporal representation learning, while \shortnameflip and \shortnamemask effectively mitigate cumulative errors and improve long-form stability. Extensive experiments demonstrate that \shortname not only achieves state-of-the-art gFVD (88.6) and VBench (81.7) scores but also enables 13× faster inference compared to existing AR baselines. These findings highlight autoregressive modeling as a practical and powerful alternative to diffusion-based approaches, paving the way toward efficient, large-scale video generation.

%% file: sec/X_suppl.tex
\clearpage
\appendix
\setcounter{page}{1}

\setcounter{table}{0}
\renewcommand{\thetable}{A\arabic{table}}
\setcounter{figure}{0}
\renewcommand{\thefigure}{A\arabic{figure}}

\maketitlesupplementary

\begin{figure*}[t]
\centering
{
\includegraphics[width=2.1\columnwidth]{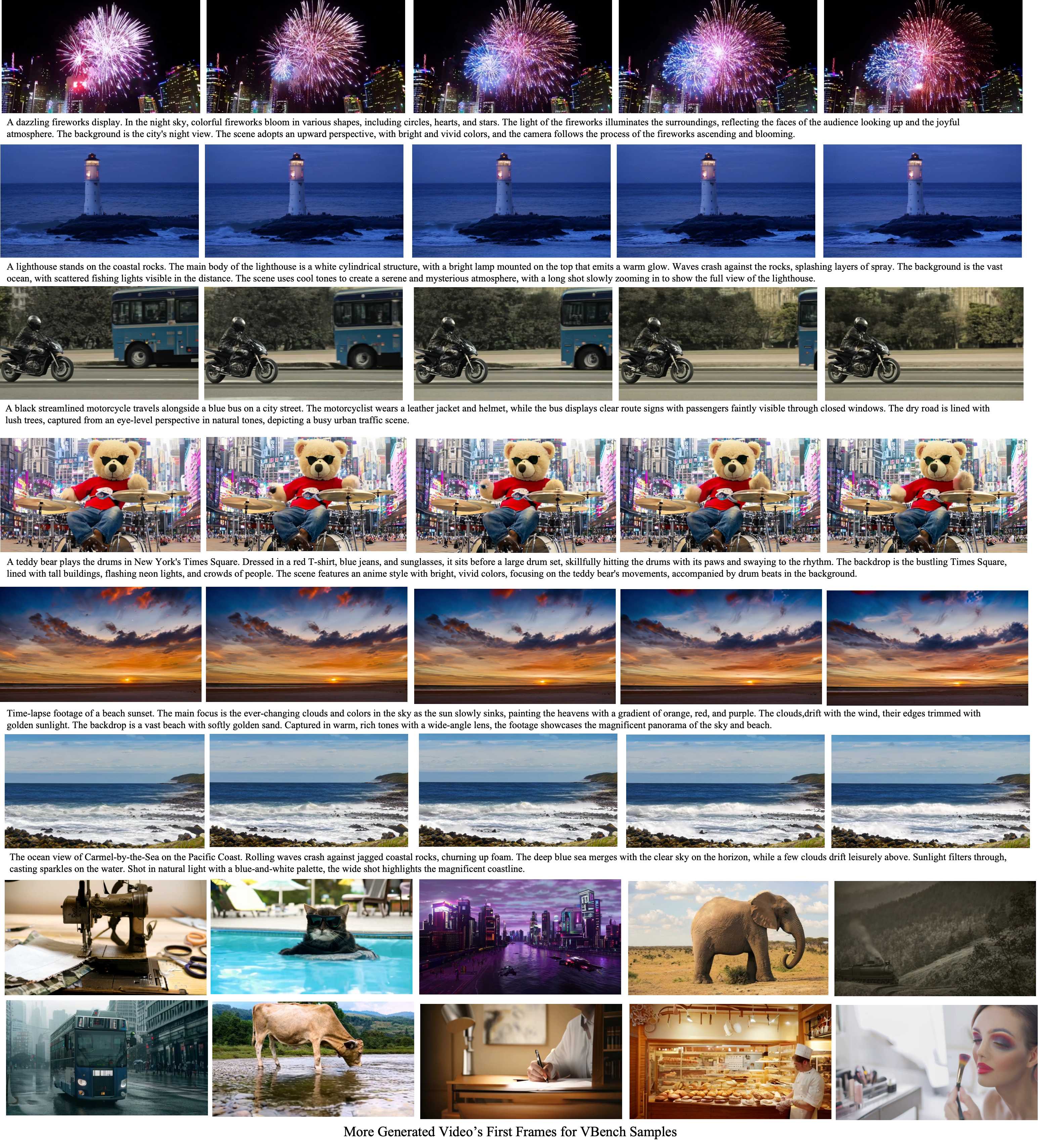} }
\vspace{-3mm}
\caption{
Visualization of more generated videos of our VideoAR-4B model.}
\label{suppl_show_t2v}
\vspace{-0.2cm}
\end{figure*}


\section{Implementation Details.}

\textbf{Model Architecture.} Our video generation framework comprises a video tokenizer and an autoregressive Transformer. The tokenizer, which maps video clips into a discrete latent space, is spatially initialized with the pre-trained weights of the publicly available Infinity image model. It operates with a latent dimension of $d=48$ and a temporal subsampling ratio of $\tau=4$. The core generative module is an autoregressive Transformer whose scale follows the architectural principles of the LLaMA family~\cite{touvron2023llama}. \shortname-L/XL employs 24/32 layers Transformer with a hidden dimension of 1536/2048 and 12/16 heads. For large-scale experiments, we further introduce \shortname-4B with 36 layers, 2560-dim and 32 heads.


\textbf{Training Configuration.} We first fine-tune the video tokenizer on the \datasetUCF dataset for 2000 epochs with a batch size of 128. Subsequently, the autoregressive Transformer is trained using the AdamW optimizer ($\beta_1=0.9, \beta_2=0.95$) with a cosine learning rate schedule of $1\times10^{-4}$ and a weight decay of $0.05$. We employ several key strategies during this phase: (i) \shortnameflip (\cref{eq:time-flip}) with parameters $p_{\min}=0$, $p_{\max}=0.25$, and $\delta=0.01$; and (ii) \shortnamemask (\cref{eq:rand-mask-attn}). For experiments on our large-scale internal dataset, we use the same hyperparameter settings while extending the training duration to account for the increased data complexity. To optimize computational efficiency, we utilize mixed-precision training and gradient checkpointing.

\textbf{Inference.} During inference, we adopt different temporal–spatial CFG schedules for class-conditional \datasetUCF and real-world text-to-video generation. For \datasetUCF, to balance sampling diversity and visual quality, we gradually increase the initial CFG strength in each frame's first scale from $\gamma = 1 \to 5$, followed by a linear increase within each frame up to $\gamma = 10$. For real-world text-to-video generation, we reduce the CFG from $\gamma = 5 \to 3$ along each temporal-spatial dimension together to maintain stronger spatial consistency in the generated videos.

\section{Visualization of Generated Videos}
In this section, we present additional generated samples of our VideoAR-4B on VBench benchmarks. As illustrated in Figure~\ref{suppl_show_t2v}, VideoAR-4B demonstrates the capability to generate high-fidelity and temporally consistent videos across diverse domains. Specifically, it can maintain object consistency in high-dynamic scenes (e.g., fireworks displays, drum-playing performances), synthesize high-aesthetic natural scenery (e.g., sunsets, the Pacific Coast), precisely adhere to semantic guidance for multi-object scenarios, and generate stylized content such as Cyberpunk-themed videos and imaginative visuals.
\section{Limitations and Future Work}
Through extensive experiments, we identify three primary limitations of the current model:
\begin{itemize}
\item \textbf{Limited Resolution and FPS}.
Our current VideoAR-4B generates videos at a resolution of 384×672 and 8 frames per second (FPS), which is insufficient for commercial applications—where standard specifications typically include 24 FPS and 720P resolution. This constraint stems from limited computational resources during training, which restrict the maximum sequence length. Additionally, the adoption of a full autoregressive (VAR) attention mask results in high computational overhead. In future work, we will extend the training sequence length and explore sparser attention mechanisms to enable high-resolution, fluid video generation.
\item \textbf{Drifting Issues in High-Dynamic Scenarios}.
During experimentation, we observe that VideoAR-4B tends to produce drifted motions for high-dynamic scenes (e.g., complex human movements). This phenomenon arises from the error-propagation inherent to autoregressive models. To address this, future research will enhance the model’s performance by integrating iterative inference-time roll-outs and reinforcement learning algorithms.
\end{itemize}

%% file: main.bib
@String(IJCV = {Int. J. Comput. Vis.})

@String(IJCV  = {IJCV})

@article{agarwal2025cosmos,
  title={Cosmos world foundation model platform for physical ai},
  author={Agarwal, Niket and Ali, Arslan and Bala, Maciej and Balaji, Yogesh and Barker, Erik and Cai, Tiffany and Chattopadhyay, Prithvijit and Chen, Yongxin and Cui, Yin and Ding, Yifan and others},
  journal={arXiv preprint arXiv:2501.03575},
  year={2025}
}

@article{wang2024loong,
  title={Loong: Generating minute-level long videos with autoregressive language models},
  author={Wang, Yuqing and Xiong, Tianwei and Zhou, Daquan and Lin, Zhijie and Zhao, Yang and Kang, Bingyi and Feng, Jiashi and Liu, Xihui},
  journal={arXiv preprint arXiv:2410.02757},
  year={2024}
}

@article{yu2025videomar,
  title={VideoMAR: Autoregressive Video Generatio with Continuous Tokens},
  author={Yu, Hu and Gong, Biao and Yuan, Hangjie and Zheng, DanDan and Chai, Weilong and Chen, Jingdong and Zheng, Kecheng and Zhao, Feng},
  journal={arXiv preprint arXiv:2506.14168},
  year={2025}
}

@article{wang2024emu3,
  title={Emu3: Next-token prediction is all you need},
  author={Wang, Xinlong and Zhang, Xiaosong and Luo, Zhengxiong and Sun, Quan and Cui, Yufeng and Wang, Jinsheng and Zhang, Fan and Wang, Yueze and Li, Zhen and Yu, Qiying and others},
  journal={arXiv preprint arXiv:2409.18869},
  year={2024}
}

@article{kong2024hunyuanvideo,
  title={Hunyuanvideo: A systematic framework for large video generative models},
  author={Kong, Weijie and Tian, Qi and Zhang, Zijian and Min, Rox and Dai, Zuozhuo and Zhou, Jin and Xiong, Jiangfeng and Li, Xin and Wu, Bo and Zhang, Jianwei and others},
  journal={arXiv preprint arXiv:2412.03603},
  year={2024}
}

@software{gen3,
  title        = {Runway Gen-3},
  author       = {Runway},
  year         = {2024},
  url          = {https://runwayml.com/research/gen3},
  note         = {Closed-source video generation model}
}

@article{Ma2025StepVideoT2VTR,
  title={Step-Video-T2V Technical Report: The Practice, Challenges, and Future of Video Foundation Model},
  author={Guoqing Ma and Haoyang Huang and Kun Yan and Liangyu Chen and Nan Duan and Sheng-Siang Yin and Changyi Wan and Ranchen Ming and Xiaoniu Song and Xing Chen and Yu Zhou and Deshan Sun and Deyu Zhou and Jian Zhou and Kaijun Tan and Kang An and Mei Chen and Wei Ji and Qiling Wu and Wenzheng Sun and Xin Han and Yana Wei and Zheng Ge and Aojie Li and Bin Wang and Bizhu Huang and Bo Wang and Brian Li and Changxing Miao and Chen Xu and Chenfei Wu and Chenguang Yu and Da Shi and Dingyuan Hu and Enle Liu and Gang Yu and Gege Yang and Guanzhe Huang and Gulin Yan and Hai-bo Feng and Hao Nie and Hao Jia and Hanpeng Hu and Hanqi Chen and Haolong Yan and Heng Wang and Hong-Wei Guo and Huilin Xiong and Hui Xiong and Jiahao Gong and Jianchang Wu and Jiao Wu and Jie Wu and Jie Yang and Jiashuai Liu and Jiashuo Li and Jingyang Zhang and Jun-Nan Guo and Junzhe Lin and Kai-hua Li and Lei Liu and Lei Xia and Liang Zhao and Liguo Tan and Liwen Huang and Li-Li Shi and Ming Li and Mingliang Li and Muhua Cheng and Na Wang and Qiao-Li Chen and Qi He and Qi Liang and Quan Sun and Ran Sun and Rui Wang and Shaoliang Pang and Shi-kui Yang and Si-Ye Liu and Siqi Liu and Shu-Guang Gao and Tiancheng Cao and Tianyu Wang and Weipeng Ming and Wenqing He and Xuefeng Zhao and Xuelin Zhang and Xi Zeng and Xiaojian Liu and Xuan Yang and Ya‐Nan Dai and Yanbo Yu and Yang Li and Yin-Yong Deng and Yingming Wang and Yilei Wang and Yuanwei Lu and Yu Chen and Yu Luo and Yu Luo},
  journal={ArXiv},
  year={2025},
  volume={abs/2502.10248},
  url={https://api.semanticscholar.org/CorpusID:276395073}
}

@misc{chen2024videocrafter2,
      title={VideoCrafter2: Overcoming Data Limitations for High-Quality Video Diffusion Models}, 
      author={Haoxin Chen and Yong Zhang and Xiaodong Cun and Menghan Xia and Xintao Wang and Chao Weng and Ying Shan},
      year={2024},
      eprint={2401.09047},
      archivePrefix={arXiv},
      primaryClass={cs.CV}
}

@article{wang2023lavie,
  title={LAVIE: High-Quality Video Generation with Cascaded Latent Diffusion Models},
  author={Wang, Yaohui and Chen, Xinyuan and Ma, Xin and Zhou, Shangchen and Huang, Ziqi and Wang, Yi and Yang, Ceyuan and He, Yinan and Yu, Jiashuo and Yang, Peiqing and others},
  journal={IJCV},
  year={2024}
}

@software{kling,
  title        = {Kling},
  author       = {Kuaishou Technology},
  year         = {2024},
  url          = {https://kling.kuaishou.com}
}

@Article{VAR,
      title={Visual Autoregressive Modeling: Scalable Image Generation via Next-Scale Prediction}, 
      author={Keyu Tian and Yi Jiang and Zehuan Yuan and Bingyue Peng and Liwei Wang},
      year={2024},
      eprint={2404.02905},
      archivePrefix={arXiv},
      primaryClass={cs.CV}
}

@inproceedings{van2017vqvae,
 author = {van den Oord, Aaron and Vinyals, Oriol and kavukcuoglu, koray},
 booktitle = {Advances in Neural Information Processing Systems},
 editor = {I. Guyon and U. Von Luxburg and S. Bengio and H. Wallach and R. Fergus and S. Vishwanathan and R. Garnett},
 pages = {},
 publisher = {Curran Associates, Inc.},
 title = {Neural Discrete Representation Learning},
 url = {https://proceedings.neurips.cc/paper_files/paper/2017/file/7a98af17e63a0ac09ce2e96d03992fbc-Paper.pdf},
 volume = {30},
 year = {2017}
}

@article{lipman2022flow,
  title={Flow matching for generative modeling},
  author={Lipman, Yaron and Chen, Ricky TQ and Ben-Hamu, Heli and Nickel, Maximilian and Le, Matt},
  journal={arXiv preprint arXiv:2210.02747},
  year={2022}
}

@inproceedings{huang2024vbench,
  title={Vbench: Comprehensive benchmark suite for video generative models},
  author={Huang, Ziqi and He, Yinan and Yu, Jiashuo and Zhang, Fan and Si, Chenyang and Jiang, Yuming and Zhang, Yuanhan and Wu, Tianxing and Jin, Qingyang and Chanpaisit, Nattapol and others},
  booktitle={Proceedings of the IEEE/CVF Conference on Computer Vision and Pattern Recognition},
  pages={21807--21818},
  year={2024}
}

@inproceedings{yu2023magvit,
  title={Magvit: Masked generative video transformer},
  author={Yu, Lijun and Cheng, Yong and Sohn, Kihyuk and Lezama, Jos{\'e} and Zhang, Han and Chang, Huiwen and Hauptmann, Alexander G and Yang, Ming-Hsuan and Hao, Yuan and Essa, Irfan and others},
  booktitle={Proceedings of the IEEE/CVF Conference on Computer Vision and Pattern Recognition},
  pages={10459--10469},
  year={2023}
}

@article{yu2023language,
  title={Language Model Beats Diffusion--Tokenizer is Key to Visual Generation},
  author={Yu, Lijun and Lezama, Jos{\'e} and Gundavarapu, Nitesh B and Versari, Luca and Sohn, Kihyuk and Minnen, David and Cheng, Yong and Birodkar, Vighnesh and Gupta, Agrim and Gu, Xiuye and others},
  journal={arXiv preprint arXiv:2310.05737},
  year={2023}
}

@inproceedings{han2025infinity,
  title={Infinity: Scaling bitwise autoregressive modeling for high-resolution image synthesis},
  author={Han, Jian and Liu, Jinlai and Jiang, Yi and Yan, Bin and Zhang, Yuqi and Yuan, Zehuan and Peng, Bingyue and Liu, Xiaobing},
  booktitle={Proceedings of the Computer Vision and Pattern Recognition Conference},
  pages={15733--15744},
  year={2025}
}

@inproceedings{tseng2021regularizing,
  title={Regularizing generative adversarial networks under limited data},
  author={Tseng, Hung-Yu and Jiang, Lu and Liu, Ce and Yang, Ming-Hsuan and Yang, Weilong},
  booktitle={Proceedings of the IEEE/CVF conference on computer vision and pattern recognition},
  pages={7921--7931},
  year={2021}
}

@article{su2024roformer,
  title={Roformer: Enhanced transformer with rotary position embedding},
  author={Su, Jianlin and Ahmed, Murtadha and Lu, Yu and Pan, Shengfeng and Bo, Wen and Liu, Yunfeng},
  journal={Neurocomputing},
  volume={568},
  pages={127063},
  year={2024},
  publisher={Elsevier}
}

@article{soomro2012ucf101,
  title={Ucf101: A dataset of 101 human actions classes from videos in the wild},
  author={Soomro, Khurram and Zamir, Amir Roshan and Shah, Mubarak},
  journal={arXiv preprint arXiv:1212.0402},
  year={2012}
}

@article{unterthiner2019fvd,
  title={FVD: A new metric for video generation},
  author={Unterthiner, Thomas and Van Steenkiste, Sjoerd and Kurach, Karol and Marinier, Rapha{\"e}l and Michalski, Marcin and Gelly, Sylvain},
  year={2019}
}

@article{wang2024omnitokenizer,
  title={Omnitokenizer: A joint image-video tokenizer for visual generation},
  author={Wang, Junke and Jiang, Yi and Yuan, Zehuan and Peng, Bingyue and Wu, Zuxuan and Jiang, Yu-Gang},
  journal={Advances in Neural Information Processing Systems},
  volume={37},
  pages={28281--28295},
  year={2024}
}

@inproceedings{wang2025parallelized,
  title={Parallelized autoregressive visual generation},
  author={Wang, Yuqing and Ren, Shuhuai and Lin, Zhijie and Han, Yujin and Guo, Haoyuan and Yang, Zhenheng and Zou, Difan and Feng, Jiashi and Liu, Xihui},
  booktitle={Proceedings of the Computer Vision and Pattern Recognition Conference},
  pages={12955--12965},
  year={2025}
}

@inproceedings{ge2022long,
  title={Long video generation with time-agnostic vqgan and time-sensitive transformer},
  author={Ge, Songwei and Hayes, Thomas and Yang, Harry and Yin, Xi and Pang, Guan and Jacobs, David and Huang, Jia-Bin and Parikh, Devi},
  booktitle={European Conference on Computer Vision},
  pages={102--118},
  year={2022},
  organization={Springer}
}

@article{hong2022cogvideo,
  title={Cogvideo: Large-scale pretraining for text-to-video generation via transformers},
  author={Hong, Wenyi and Ding, Ming and Zheng, Wendi and Liu, Xinghan and Tang, Jie},
  journal={arXiv preprint arXiv:2205.15868},
  year={2022}
}

@software{Veo3,
  title        = {Veo 3},
  author       = {{Google DeepMind}},
  year         = {2025},
  url          = {https://deepmind.google/technologies/veo/},
  note         = {Closed-source video generation model.}
}

@software{Sora2,
  title        = {Sora 2},
  author       = {OpenAI},
  year         = {2025},
  url          = {https://openai.com/research/sora},
  note         = {Closed-source video generation model.}
}

@article{wan2025,
      title={Wan: Open and Advanced Large-Scale Video Generative Models}, 
      author={Team Wan and Ang Wang and Baole Ai and Bin Wen and Chaojie Mao and Chen-Wei Xie and Di Chen and Feiwu Yu and Haiming Zhao and Jianxiao Yang and Jianyuan Zeng and Jiayu Wang and Jingfeng Zhang and Jingren Zhou and Jinkai Wang and Jixuan Chen and Kai Zhu and Kang Zhao and Keyu Yan and Lianghua Huang and Mengyang Feng and Ningyi Zhang and Pandeng Li and Pingyu Wu and Ruihang Chu and Ruili Feng and Shiwei Zhang and Siyang Sun and Tao Fang and Tianxing Wang and Tianyi Gui and Tingyu Weng and Tong Shen and Wei Lin and Wei Wang and Wei Wang and Wenmeng Zhou and Wente Wang and Wenting Shen and Wenyuan Yu and Xianzhong Shi and Xiaoming Huang and Xin Xu and Yan Kou and Yangyu Lv and Yifei Li and Yijing Liu and Yiming Wang and Yingya Zhang and Yitong Huang and Yong Li and You Wu and Yu Liu and Yulin Pan and Yun Zheng and Yuntao Hong and Yupeng Shi and Yutong Feng and Zeyinzi Jiang and Zhen Han and Zhi-Fan Wu and Ziyu Liu},
      journal = {arXiv preprint arXiv:2503.20314},
      year={2025}
}

@article{touvron2023llama,
  title={Llama: Open and efficient foundation language models},
  author={Touvron, Hugo and Lavril, Thibaut and Izacard, Gautier and Martinet, Xavier and Lachaux, Marie-Anne and Lacroix, Timothy and Rozi{\`e}re, Baptiste and Goyal, Naman and Hambro, Eric and Azhar, Faisal and Rodriguez, Aurelien and Joulin, Armand and Grave, Edouard and Lample, Guillaume},
  journal={arXiv preprint arXiv:2302.13971},
  year={2023}
}

@inproceedings{rombach2022high,
  title={High-resolution image synthesis with latent diffusion models},
  author={Rombach, Robin and Blattmann, Andreas and Lorenz, Dominik and Esser, Patrick and Ommer, Bj{\"o}rn},
  booktitle={Proceedings of the IEEE/CVF conference on computer vision and pattern recognition},
  pages={10684--10695},
  year={2022}
}

@article{sun2024autoregressive,
  title={Autoregressive Model Beats Diffusion: Llama for Scalable Image Generation},
  author={Sun, Peize and Jiang, Yi and Chen, Shoufa and Zhang, Shilong and Peng, Bingyue and Luo, Ping and Yuan, Zehuan},
  journal={arXiv preprint arXiv:2406.06525},
  year={2024}
}

@article{deng2025bagel,
  title   = {Emerging Properties in Unified Multimodal Pretraining},
  author  = {Deng, Chaorui and Zhu, Deyao and Li, Kunchang and Gou, Chenhui and Li, Feng and Wang, Zeyu and Zhong, Shu and Yu, Weihao and Nie, Xiaonan and Song, Ziang and Shi, Guang and Fan, Haoqi},
  journal = {arXiv preprint arXiv:2505.14683},
  year    = {2025}
}

@misc{ai2025magi1autoregressivevideogeneration,
      title={MAGI-1: Autoregressive Video Generation at Scale},
      author={Sand. ai and Hansi Teng and Hongyu Jia and Lei Sun and Lingzhi Li and Maolin Li and Mingqiu Tang and Shuai Han and Tianning Zhang and W. Q. Zhang and Weifeng Luo and Xiaoyang Kang and Yuchen Sun and Yue Cao and Yunpeng Huang and Yutong Lin and Yuxin Fang and Zewei Tao and Zheng Zhang and Zhongshu Wang and Zixun Liu and Dai Shi and Guoli Su and Hanwen Sun and Hong Pan and Jie Wang and Jiexin Sheng and Min Cui and Min Hu and Ming Yan and Shucheng Yin and Siran Zhang and Tingting Liu and Xianping Yin and Xiaoyu Yang and Xin Song and Xuan Hu and Yankai Zhang and Yuqiao Li},
      year={2025},
      eprint={2505.13211},
      archivePrefix={arXiv},
      primaryClass={cs.CV},
      url={https://arxiv.org/abs/2505.13211},
}

@article{liu2025infinitystar,
  title={Infinitystar: Unified spacetime autoregressive modeling for visual generation},
  author={Liu, Jinlai and Han, Jian and Yan, Bin and Wu, Hui and Zhu, Fengda and Wang, Xing and Jiang, Yi and Peng, Bingyue and Yuan, Zehuan},
  journal={arXiv preprint arXiv:2511.04675},
  year={2025}
}
